\documentclass{article}

\usepackage{microtype}
\usepackage{graphicx}
\usepackage{subcaption}
\usepackage{booktabs}
\usepackage{hyperref}
\usepackage{booktabs,array,longtable}

\usepackage[preprint]{icml2027}
\usepackage{amsmath}
\usepackage{amssymb}
\usepackage{mathtools}
\usepackage{amsthm}
\usepackage{bm}
\usepackage[capitalize,noabbrev]{cleveref}
\hypersetup{pdfsubject={Preprint}}

\icmltitlerunning{Two-Stage Post-Training for Flow-Based Embodied Foundation Models}

\begin{document}

\twocolumn[
  \icmltitle{Two-Stage Post Training for Flow Based Embodied Models via Multi-Teacher On Policy Distillation}

  \begin{icmlauthorlist}
    \icmlauthor{Jiaming Zhang}{inst1}
    \icmlauthor{Yao Lyu}{inst2}
    \icmlauthor{Haoning Wang}{inst1}
    \icmlauthor{Yongjun Chen}{inst1}
    \icmlauthor{Liping Zhang}{inst1}
    \icmlauthor{Shengbo Eben Li}{inst3}
  \end{icmlauthorlist}
  \icmlaffiliation{inst1}{Department of Mathematical Sciences, Tsinghua University, Beijing, 100084, China}
  \icmlaffiliation{inst2}{School of Vehicle and Mobility, Tsinghua University, Beijing, 100084, China}
  \icmlaffiliation{inst3}{School of Vehicle and Mobility \& College of AI, Tsinghua University, Beijing, 100084, China}
  \icmlcorrespondingauthor{Liping Zhang}{lipingzhang@tsinghua.edu.cn}
  \icmlcorrespondingauthor{Shengbo Eben Li}{lishbo@tsinghua.edu.cn}

  \icmlkeywords{Embodied AI, Flow Matching, On Policy Distillation, Post Training, Reinforcement Learning}
  \vskip 0.3in ]

\printAffiliationsAndNotice{}

\begin{abstract}
Large-scale pretraining provides embodied models with general capabilities, but reliable task execution still requires post training.
In multitask settings, supervised fine-tuning (SFT) faces distribution mismatch
between training and deployment, whereas reinforcement learning (RL) is limited by sparse rewards and uneven learning progress across tasks. Multi-teacher on policy distillation (MOPD) consolidates expert capabilities through dense supervision on the model's own trajectories, but achieving performance beyond the teachers remains challenging.
We propose a two-stage post training framework for flow-based embodied models, constructing distillation objectives from the velocity regression formulation of conditional flow matching.
Stage 1 efficiently consolidates expert capabilities through MOPD.
Stage 2 combines RL with teacher distillation that weights velocity errors according to the sampling noise, where distillation is restricted to failed trajectories to supplement sparse rewards with dense guidance without
directly constraining successful behaviors.
Experiments across multiple models and benchmarks show that the framework outperforms joint SFT and exceeds the corresponding teachers' aggregate success rates. Sample efficiency comparisons show higher success rates than direct RL with fewer environment interactions.
\end{abstract}

\section{Introduction}
\label{sec:introduction}
Large-scale pretraining enables embodied models to acquire generalizable representations and skills for robot control across diverse tasks and environments. For example, vision-language-action (VLA) models generate actions from visual observations, language instructions and proprioceptive states \citep{pi0}, while world action models (WAMs) jointly predict future observations and actions \citep{dreamzero}. Despite their broad capabilities, these models often require post training to achieve reliable performance on target tasks. Maintaining a separate policy for each task incurs storage and maintenance costs, along with the overhead of switching between models during deployment \citep{nestedvla, robion}. These costs increase as the task set grows, motivating post training methods that improve multitask performance within a single shared policy.

Supervised fine-tuning (SFT) and reinforcement learning (RL) are two common approaches to post training. SFT trains models on expert demonstrations, using observations and task conditions as inputs and deriving supervision from expert actions \citep{pi0}. RL instead uses trajectories collected through the model’s own interaction with the environment to optimize expected cumulative reward \citep{pirl}. These approaches have complementary strengths and limitations. SFT provides dense supervision that supports efficient task learning, but its effectiveness depends on demonstration quality and coverage. During execution, the model may encounter states poorly represented in the demonstrations, limiting its ability to generalize beyond the demonstrated behavior \citep{dagger}. RL can mitigate this distribution mismatch by learning from states visited by the model itself. However, sparse rewards make it difficult to determine how individual actions contribute to task outcomes, complicating credit assignment and reducing learning efficiency \citep{vlarl}. In multitask settings, SFT requires demonstrations covering a wider range of tasks, increasing data collection costs \citep{droid}, while uneven learning progress and imbalanced reward feedback across tasks complicate joint optimization for RL \citep{mtopt}. Combining online interaction with dense expert supervision offers a natural way to retain these complementary advantages, allowing models to learn from expert guidance on states they visit during execution.

On policy distillation (OPD) provides such guidance through trained teacher models that supervise learning on trajectories generated by the model being optimized. In large language models (LLMs), OPD trains the model to match the teacher’s next-token predictive distribution at each position in its own generated sequences. Multi-teacher on policy distillation (MOPD) extends this approach by selecting specialized teachers for different tasks and consolidating their capabilities into a shared model. It has become an important paradigm for capability integration during LLM post training \citep{mimo,kimi,deepseek}. By providing dense supervision throughout generated trajectories, OPD can improve sample efficiency and reduce training costs compared with sparse-reward RL \citep{qwen3,vlaopd}. However, consolidating multiple experts does not guarantee that the resulting model matches each teacher on its respective tasks. Different teachers can induce conflicting updates in shared parameters, improving performance on some tasks at the expense of others \citep{roadmopd,pmopd}. Moreover, matching teacher predictions does not directly optimize task success, so reducing distillation error need not improve closed-loop performance. These limitations motivate combining environmental feedback with teacher supervision to further improve task performance after consolidating expert capabilities.

For flow-based embodied models, teacher supervision must also account for the continuous action generation process. Unlike autoregressive models that generate discrete tokens, these models generate continuous action chunks by integrating a learned velocity field \citep{flowmatching}. DiffusionOPD introduces stochasticity through SDE sampling to enable teacher supervision via closed-form KL minimization between Gaussian transitions \citep{diffusionopd}. Recent work extends this approach to flow-based VLAs, consolidating expert capabilities through teacher supervision on noisy actions along the denoising trajectories \citep{flowmopd,vlamopd}. Building on these advances, we seek to improve task performance beyond capability consolidation. Environmental rewards guide learning toward task success, but sparse feedback provides limited guidance for individual actions. Meanwhile, applying teacher supervision to all trajectories may unnecessarily constrain successful behaviors learned by the model. We therefore propose a two-stage post training framework that combines multi-teacher on policy distillation with policy optimization driven by environmental rewards. The distillation objectives in both stages follow the velocity regression formulation of native conditional flow matching training. The first stage uses MOPD to efficiently integrate expert capabilities. The second stage combines RL with teacher supervision restricted to failed trajectories, providing dense guidance alongside environmental rewards without directly constraining successful behaviors.

Our main contributions are as follows:
\begin{itemize}
    \item We propose a two-stage post training framework for flow-based embodied models. It combines efficient expert capability consolidation through MOPD with further task improvement through RL and selective teacher guidance, enabling a shared model to acquire and refine multitask capabilities.

    \item We formulate distillation objectives based on the velocity regression principle of native conditional flow matching. The first stage matches teacher velocity predictions on the model's own noisy actions. In the second stage, we adapt this objective to the transition dynamics and noise scale of stochastic flow sampling and combine it with RL. Teacher supervision is restricted to failed trajectories, providing dense guidance alongside sparse rewards without directly constraining successful behaviors.

    \item Experiments across multiple embodied models and benchmarks demonstrate improvements in multitask success rates and sample efficiency. 
\end{itemize}

\section{Related Work}
\label{sec:related-work}

\subsection{On Policy Distillation in LLM Post-Training}
\label{sec:related-llm}

OPD matches teacher and student next-token distributions on sequences generated by the student, providing dense supervision on the contexts encountered during generation \citep{gkd,minillm}.
It has become an important paradigm in LLM post training. For integrating specialized capabilities, DeepSeek-V4 minimizes reverse KL over the full vocabulary \citep{deepseek}, while Kimi K3 selects teachers by task domain and reasoning effort \citep{kimi}. MiMo-V2-Flash combines token-level teacher signals with outcome rewards \citep{mimo}. OPD is also used to recover capabilities across training stages. GLM-5 uses earlier SFT and RL checkpoints as teachers to recover capabilities lost during subsequent training \citep{glm5}. Nemotron-Cascade 2 integrates multidomain OPD into its cascade
RL pipeline, combining the strengths of intermediate checkpoints before continuing RL \citep{nemotroncascade2}. OPD can also improve training efficiency. In a comparison from the same 8B checkpoint, Qwen3 achieves better math and coding performance with OPD than with RL,
using about one-tenth of the GPU hours \citep{qwen3}. Nemotron-Cascade 2 likewise reports comparable or better scores on two ArenaHard v2 evaluations with fewer training steps than RLHF \citep{nemotroncascade2}.

\subsection{Reinforcement Learning for Embodied Models}
\label{sec:related-rl}

RL is increasingly used for embodied post training to improve task performance through environment interaction. iRe-VLA alternates RL with supervised learning to balance exploration and training stability \citep{irevla}. SimpleVLA-RL demonstrates scalable VLA fine-tuning through parallel environment interaction and enhanced exploration under binary task rewards \citep{simplevlarl}.

For flow-based models, the lack of tractable action likelihoods poses a central challenge for policy gradient optimization. ReinFlow uses a learned noise model to construct stochastic transitions with tractable probabilities \citep{reinflow}. $\pi_{\mathrm{RL}}$ studies this formulation alongside
ODE-to-SDE conversion for VLA fine-tuning \citep{pirl}. Flow policy optimization (FPO) instead derives a likelihood-ratio surrogate from conditional flow matching losses \citep{flowmatchingpg}.
Lyu et al.\ adopt a related surrogate to optimize a latent action policy with a frozen VLA decoder \citep{lyufpo}. Value estimates provide another source of guidance for policy improvement. Q-VGM uses action-value gradients to construct velocity targets at late denoising steps, updating the VLA action expert through local velocity matching \citep{qvgm}. To reduce the cost of adapting pretrained generative models, DSRL optimizes a policy over input noise to guide action generation while keeping the base model frozen \citep{dsrl}.

\subsection{On Policy Distillation for Continuous Generative Models}
\label{sec:related-continuous}

In continuous generative models, OPD incorporates teacher supervision into iterative denoising by querying teachers along
the student's sampling trajectories. DiffusionOPD extends the conditional distribution matching objective of autoregressive OPD to Markov chains over continuous states.  It directly minimizes closed-form reverse KL between Gaussian
transitions with shared covariance under SDE sampling and uses one-step transition matching under deterministic sampling
\citep{diffusionopd}. DanceOPD represents expert capabilities as velocity fields and routes each sample to a single field for supervision at a low-noise point on the student trajectory \citep{danceopd}. FlowCTS compares teacher and student continuations from the same point on a student trajectory. An upper bound on their discrepancy yields a temporally weighted velocity matching objective that supports both single-point and multistep supervision \citep{flowcts}.

Similar approaches have been applied to embodied models. Ni et al.\ use teacher velocity supervision to consolidate
experts from LIBERO, CALVIN, and Meta-World into a single VLA \citep{flowmopd}. VLA-MOPD adopts a similar approach to consolidate expertise across suites or tasks within individual benchmarks, with
additional evaluation on a real robot \citep{vlamopd}. REPAIR instead queries local RL specialists for candidate actions at critical task phases. During on policy refinement, it weights these candidates using relative critic scores and transfers expert behavior through a weighted flow matching objective \citep{repair}.

\section{Preliminaries}
\label{sec:preliminaries}

\subsection{Flow Matching for Action Generation}
\label{sec:flow-matching}

Flow matching generates continuous action chunks by integrating a learned velocity field from Gaussian noise \citep{pi0}. At environment step $t$, the context $c=(o,l,s)$ comprises visual observations, language instructions and proprioceptive states. An action chunk $a\in\mathbb R^{H\times d}$ contains $H$ actions of dimension $d$, with flattened dimension $D=Hd$. Using denoising time $u\in[0,1]$ from noise to action, generation follows
\begin{equation}
\begin{aligned}
\mathrm dz_u=v_\theta(z_u,u\mid c)\,\ \mathrm du,
z_0\sim\mathcal N(0,I_D).
\end{aligned}
\label{eq:prelim-flow-ode}
\end{equation}
On a grid $0=u_0<\cdots<u_J=1$, Euler integration gives
\begin{equation}
z_{j+1}
=z_j+\Delta u_j\,v_\theta(z_j,u_j\mid c),
\label{eq:prelim-euler-update}
\end{equation}
where $z_j=z_{u_j}$ and $\Delta u_j=u_{j+1}-u_j$. Training samples demonstration pairs $(c,a)\sim\mathcal D$,
noise $\epsilon\sim\mathcal N(0,I_D)$, and denoising times $u\sim p(u)$. The interpolation $z_u=(1-u)\epsilon+ua$ has target velocity $a-\epsilon$, giving the conditional flow matching (CFM) objective \citep{flowmatching}:
\begin{equation}
\mathcal L_{\mathrm{CFM}}(\theta)
=
\mathbb E_{(c,a),\epsilon,u}
\left[
\operatorname{MSE}\!\left(
v_\theta(z_u,u\mid c),a-\epsilon
\right)
\right].
\label{eq:prelim-cfm-loss}
\end{equation}

\subsection{Stochastic Sampling with Flow-SDE}
\label{sec:flow-sde-sampling}

Flow-SDE introduces stochastic exploration by adding Brownian noise and adjusting the drift, which preserves the ODE's marginal distributions in continuous time \citep{flowgrpo}.
For the linear path above and $u<1$,
\begin{equation}
\begin{aligned}
\mathrm dz_u
&=b_\theta(z_u,u\mid c)\,\mathrm du
  +g(u)\,\mathrm dW_u,\\
b_\theta(z_u,u\mid c)
&=v_\theta(z_u,u\mid c)\\
&-\frac{g(u)^2}{2(1-u)}
\bigl[z_u-u\,v_\theta(z_u,u\mid c)\bigr],
\end{aligned}
\label{eq:prelim-flow-sde}
\end{equation}
where $W_u$ is standard Brownian motion and $g(u)\geq0$ controls the noise magnitude. Discretization with independent
$\xi_j\sim\mathcal N(0,I_D)$ gives
\begin{equation}
\begin{aligned}
z_{j+1}
&=\mu_\theta^j+\sqrt{\Delta u_j}\,g(u_j)\xi_j,\\
\mu_\theta^j
&=z_j+\Delta u_j\,b_\theta(z_j,u_j\mid c),\\
\Sigma_j&=\Delta u_j\,g(u_j)^2I_D.
\end{aligned}
\label{eq:prelim-sde-discrete}
\end{equation}
For $g(u_j)>0$, the local transition density is
$p_\theta^j(z_{j+1}\mid z_j,c)
=\mathcal N(z_{j+1};\mu_\theta^j,\Sigma_j)$,
enabling likelihood-ratio computation for policy gradient
optimization \citep{pirl}.

\section{Two-Stage Post Training with Multi-Teacher Distillation}
\label{sec:framework}

\subsection{Framework Overview}
\label{sec:overview}

\begin{figure*}[t]
    \centering
    \includegraphics[width=\textwidth]{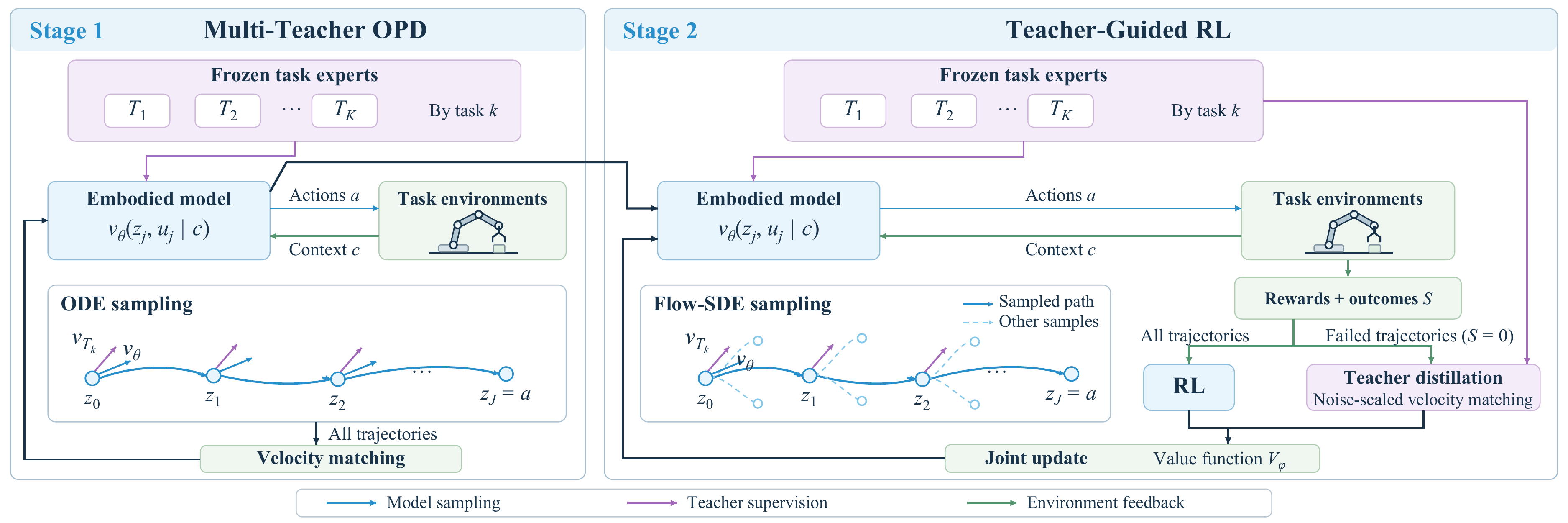}
    \caption{
    Overview of the two-stage post training framework. Stage 1 consolidates expert capabilities through MOPD; Stage 2 combines RL with teacher distillation on failed trajectories.
    }
    \label{fig:framework}
\end{figure*}

As shown in Figure~\ref{fig:framework}, we propose a two-stage post training framework for multitask flow-based embodied models. The first stage uses MOPD to efficiently consolidate expert capabilities into a shared model. Teachers provide dense supervision on observations collected through the current model's environment interactions and noisy actions along its denoising trajectories. The second stage combines RL with teacher distillation on failed trajectories, supplementing sparse rewards with dense guidance without directly constraining successful behaviors.

\subsection{Stage 1: Multi-Teacher OPD}
\label{sec:stage-one}
The first stage uses MOPD to consolidate task expertise into a shared model. We construct the distillation objective from the native CFM formulation in Equation~\eqref{eq:prelim-cfm-loss}, which learns a velocity field using intermediate actions and target velocities constructed from demonstrations and noise. We retain this velocity regression objective while changing the sources of training data and supervision signals. Training data come from the model's own
environment interactions and denoising trajectories, while the corresponding task teacher provides dense supervision through its velocity predictions. This provides dense expert guidance on the distribution encountered by the current model.

Let $\{T_k\}_{k=1}^{K}$ denote the task teachers and $v_{T_k}$ their velocity fields. Before each optimization step, we collect fresh trajectories using the current model. The samples for task group $k$ are denoted by $\mathcal B_k(\operatorname{sg}[\theta])$, where $\operatorname{sg}$ indicates that gradients do not propagate through the sampling process. With normalized task weights $\alpha_k^{(1)}$, the distillation objective is
\begin{equation}
\begin{aligned}
\mathcal L_1(\theta)
&=
\sum_{k=1}^{K}\alpha_k^{(1)}
\mathbb E_{\substack{
(c,z_{0:J})\sim\mathcal B_k(\operatorname{sg}[\theta])\\
j\sim\operatorname{Unif}\{0,\ldots,J-1\}
}}
\Bigl[
\operatorname{MSE}\!\Bigl(
\\[-2pt]
&\qquad
v_\theta(\operatorname{sg}[z_j],u_j\mid c),
\operatorname{sg}[v_{T_k}(z_j,u_j\mid c)]
\Bigr)
\Bigr],
\end{aligned}
\label{eq:stage1-mopd-loss}
\end{equation}
All teachers are pretrained task experts and remain frozen. We stop gradients through the collected denoising states, so updates act only on the model's velocity predictions at
these fixed locations. This avoids backpropagation through the denoising chain and the associated products of Jacobians~\citep{qvgm}, reducing gradient coupling and computational overhead. Teacher guidance therefore improves the local generation direction rather than changing earlier denoising steps to move the supervision points. The resulting velocity-matching objective is consistent with existing flow-based multi-teacher distillation methods~\citep{flowmopd,vlamopd}. Our formulation emphasizes its connection to native CFM training, a design principle that also guides the stochastic distillation objective in the second stage.

\subsection{Stage 2: Teacher-Guided RL}
\label{sec:stage-two}
The first stage efficiently consolidates expert capabilities, but reducing distillation error does not necessarily improve task success. Supervision from different teachers may also induce conflicting updates in shared parameters, limiting the performance achievable through distillation alone. The second stage therefore introduces RL to optimize task performance using environmental rewards. We retain teacher distillation on failed trajectories to supplement sparse rewards with dense guidance, while leaving successful behaviors free of direct teacher constraints.

We use the Flow-SDE sampler introduced in Section~\ref{sec:flow-sde-sampling} for stochastic action generation. Teacher distillation retains the velocity
regression principle of the first stage, with supervision adjusted to the stochastic transition scale. Velocity predictions determine the transition mean through the drift, while the noise magnitude determines the scale of variation around that mean. We therefore map the difference in teacher and model velocities to a difference in transition means and normalize it by the transition standard deviation. When both models are evaluated at the same denoising state under the same context, the shared state-dependent drift
terms cancel, giving
\begin{equation}
\begin{aligned}
\mu_\theta^j-\mu_{T_k}^j
&=\Delta u_j(b_\theta-b_{T_k})\\
&=\Delta u_j
\left(1+\frac{u_jg(u_j)^2}{2(1-u_j)}\right)
(v_\theta-v_{T_k}).
\end{aligned}
\label{eq:stage2-sde-mean-difference}
\end{equation}
The resulting local distillation loss is
\begin{equation}
\begin{aligned}
\ell_{\mathrm{SDE}}(\theta)
&=\operatorname{MSE}\!\left(
\frac{\mu_\theta^j}{\sqrt{\Delta u_j}\,g(u_j)},
\frac{\operatorname{sg}[\mu_{T_k}^j]}
     {\sqrt{\Delta u_j}\,g(u_j)}
\right)\\
&=\frac{\Delta u_j}{g(u_j)^2}
\left(1+\frac{u_jg(u_j)^2}{2(1-u_j)}\right)^2\\
&\quad{}\times\operatorname{MSE}\!\Bigl(
v_\theta(\operatorname{sg}[z_j],u_j\mid c),
\operatorname{sg}[v_{T_k}(z_j,u_j\mid c)]
\Bigr).
\end{aligned}
\label{eq:stage2-sde-distillation-loss}
\end{equation}
This objective actually weights velocity regression according to the sampling step size, drift correction, and noise scale.

This construction also admits an interpretation as conditional distribution matching. At the same state and context, teacher and model transitions
are Gaussian with shared covariance $\Sigma_j=\Delta u_jg(u_j)^2I_D$.
Their KL divergence therefore depends only on the normalized difference between their means, yielding
\begin{equation}
\begin{aligned}
\ell_{\mathrm{SDE}}(\theta)
&=\frac{\|\mu_\theta^j-\mu_{T_k}^j\|_2^2}
        {D\,\Delta u_jg(u_j)^2}=\frac{2}{D}
D_{\mathrm{KL}}\!\left(
p_\theta^j\middle\|p_{T_k}^j
\right).
\end{aligned}
\label{eq:stage2-sde-kl-equivalence}
\end{equation}
Thus, our velocity regression objective is proportional to the reverse KL between local transition distributions \citep{diffusionopd}.
This parallels autoregressive OPD, which matches next-token distributions at generated prefixes, while here supervision matches the next stochastic transition at collected denoising states.

Building on this local teacher supervision, we combine environmental rewards and teacher guidance in a joint training objective. Following the Flow-SDE RL formulation \citep{pirl}, we estimate advantages $\hat A_t$ using generalized advantage estimation (GAE) and optimize the model with PPO. Let $\theta_{\mathrm{old}}$ denote the fixed parameter snapshot used to collect the current batch. The probability ratio and per-sample PPO loss are
\begin{equation}
\begin{aligned}
r_\theta
&=
\frac{p_\theta^j(z_{j+1}\mid z_j,c_t)}
     {p_{\theta_{\mathrm{old}}}^j(z_{j+1}\mid z_j,c_t)},\\
\ell_{\mathrm{PPO}}(\theta)
&=-\min\!\Bigl(
r_\theta\hat A_t,\,
\operatorname{clip}(r_\theta,1-\epsilon,1+\epsilon)
\hat A_t
\Bigr),
\end{aligned}
\label{eq:stage2-ppo-loss}
\end{equation}
where $\epsilon$ is the clipping threshold. The value function $V_\phi$ is trained with a squared-error loss $\ell_V(\phi)$ against GAE-based return targets. All collected samples contribute to RL, whereas teacher distillation is restricted to trajectories that failed during collection. Let $S\in\{0,1\}$ indicate the trajectory outcome, with $S=1$ denoting success and $S=0$ denoting failure. The combined objective is
\begin{equation}
\begin{aligned}
\mathcal L_2(\theta,\phi)
&=
\mathbb E_{\substack{
(c_t,z_{0:J},j,S)\sim\\
\mathcal B(\theta_{\mathrm{old}})
}}
\left[
\ell_{\mathrm{PPO}}(\theta)+c_V\ell_V(\phi)
\right]\\
&\quad+
\beta\sum_{k=1}^{K}\alpha_k^{(2)}
\mathbb E_{\substack{
(c_t,z_{0:J},j,S)\sim\\
\mathcal B_k(\theta_{\mathrm{old}})
}}
\left[
(1-S)\ell_{\mathrm{SDE}}(\theta)
\right],
\end{aligned}
\label{eq:stage2-joint-loss}
\end{equation}
where $\mathcal B_k(\theta_{\mathrm{old}})$ contains samples from task group $k$, and $\mathcal B(\theta_{\mathrm{old}})$ pools samples across groups. The normalized weights $\alpha_k^{(2)}$ balance teacher distillation across tasks; $\beta$ and $c_V$ weight the distillation and value losses. The two-stage training procedure is summarized in Algorithm~\ref{alg:two-stage-training}.

\begin{algorithm}[t]
\caption{Two-Stage Post Training with Multi-Teacher Distillation}
\label{alg:two-stage-training}
\begin{algorithmic}[1]

\REQUIRE Model parameters $\theta$; value function $V_\phi$;
frozen teachers $\{T_k\}_{k=1}^{K}$;
task and loss weights;
learning rates $\eta_1,\eta_\theta,\eta_\phi$;
PPO update epochs $E$.
\ENSURE Post-trained model parameters $\theta$.

\STATE \textbf{Stage 1: Multi-teacher on policy distillation}
\FOR{each training round}
    \FOR{$k=1,\ldots,K$}
        \STATE $\mathcal B_k(\operatorname{sg}[\theta])
        \leftarrow
        \operatorname{Rollout}(\theta,k,\mathrm{ODE})$
    \ENDFOR
    \STATE Compute $\mathcal L_1(\theta)$ using
    Equation~\eqref{eq:stage1-mopd-loss} and update $\theta$ with one optimizer step:
    $\theta\leftarrow\theta-\eta_1\nabla_\theta
    \mathcal L_1(\theta)$.
\ENDFOR

\STATE \textbf{Stage 2: Teacher-guided reinforcement learning}
\STATE Initialize $\theta$ from the selected Stage 1 checkpoint.
\FOR{each training round}
    \STATE $\theta_{\mathrm{old}}\leftarrow\theta$.
    \FOR{$k=1,\ldots,K$}
        \STATE $\mathcal B_k(\theta_{\mathrm{old}})
        \leftarrow
        \operatorname{Rollout}
        (\theta_{\mathrm{old}},k,\text{Flow-SDE})$.
    \ENDFOR
    \STATE $\mathcal B\leftarrow
    \bigcup_{k=1}^{K}\mathcal B_k(\theta_{\mathrm{old}})$.
    \STATE Compute and detach advantages $\hat A_t$
and value targets $\hat G_t$ from $\mathcal B$ using GAE.
\STATE Construct and fix the failure mask $m=1-S$
on $\mathcal B$.
    \FOR{$e=1,\ldots,E$}
    \FOR{each global batch $\mathcal G$ from $\mathcal B$}
        \STATE Compute $\widehat{\mathcal L}_2(\theta,\phi)$
on $\mathcal G$ using Equation~\eqref{eq:stage2-joint-loss}
and update
$\theta\leftarrow\theta-\eta_\theta\nabla_\theta
\widehat{\mathcal L}_2$,
$\phi\leftarrow\phi-\eta_\phi\nabla_\phi
\widehat{\mathcal L}_2$.
    \ENDFOR
\ENDFOR
\ENDFOR
\end{algorithmic}
\end{algorithm}

\section{Experiments}
\label{sec:experiments}

\subsection{Experimental Setup}
\label{sec:experimental-setup}

We conduct experiments with three flow-based embodied models, GR00T N1.5 ~\citep{grootn1}, QwenVLA ~\citep{qwenvla}, and $\pi_{0.5}$ ~\citep{pi05}, across LIBERO ~\citep{libero}, RoboCasa Kitchen ~\citep{robocasa}, and RoboCasa GR1 ~\citep{grootn1}. LIBERO comprises four suites: Spatial, Object, Goal, and Long. For RoboCasa, we group tasks by similarity. Kitchen comprises PnP, Articulated, and Appliance--Coffee, with eight tasks each. These cover pick-and-place, door and
drawer manipulation, and appliance operation. GR1 comprises Articulated, Tableware, and Storage, with six, nine, and nine tasks. These cover placing objects and closing articulated parts, transfers involving tableware and cookware, and storage. Within a benchmark, we train one shared model to handle all suites or task groups. Task partitions for RoboCasa and key training hyperparameters are provided in Appendix~\ref{app:implementation}.

Each model is initialized from a multitask SFT checkpoint for the corresponding benchmark. Each task group has a teacher model trained through SFT or RL on its tasks. Teachers remain frozen during post training and are
selected according to the task group. We freeze the vision-language backbone and fine-tune the action generation module, including the action expert and associated input and output projection layers. In the second stage, we also train a value network. Training uses eight NVIDIA H100 GPUs with 80~GB of memory each.

We use each benchmark's official environments, task initialization procedures, and success criteria. To reduce evaluation costs during training, we evaluate every 10 rounds using 20 episodes per task. These evaluations are used to plot
learning curves and select the checkpoint with the highest mean success rate across tasks. For the main results, we evaluate the selected checkpoint over 50 episodes per task, following prior studies ~\citep{pirl,robocasa,dial}. Overall success rates are computed by averaging task success rates with equal weight.

\subsection{Multitask Performance}
\label{sec:main-results}

We evaluate the proposed post training framework across models and benchmarks, examining multitask performance and the gains from each training stage. Tables~\ref{tab:main-libero} and~\ref{tab:main-robocasa} summarize the results for the five experimental settings. Initialization denotes the multitask SFT checkpoint used for post training. Stage 1 and Stage 1 + Stage 2 denote
the intermediate and final models. The Teachers rows report expert performance for each task group, and the SFT rows provide reference results from prior studies or official repositories.

\begin{table}[t]
\centering
\caption{Success rates (\%) on LIBERO.}
\label{tab:main-libero}
\fontsize{8}{9.2}\selectfont
\setlength{\tabcolsep}{2pt}
\renewcommand{\arraystretch}{1.0}
\setlength{\aboverulesep}{1.2pt}
\setlength{\belowrulesep}{1.2pt}
\begin{tabular*}{\columnwidth}{@{\extracolsep{\fill}}lrrrrr@{}}
\toprule
Method & Spatial & Object & Goal & Long & \textbf{Avg.} \\
\midrule
\multicolumn{6}{@{}l}{\textbf{GR00T N1.5}} \\
Initialization & 81.40 & 93.80 & 88.80 & 64.40 & 82.10 \\
Teachers & 98.00 & 98.60 & 95.60 & \underline{96.80} & \underline{97.25} \\
\shortstack[l]{SFT \citep{koo2026hamlet}} & \underline{98.20} & \underline{99.40} & \underline{97.20} & 87.80 & 95.65 \\
Stage 1 & 96.60 & 97.40 & 93.00 & 93.60 & 95.15 \\
Stage 1 + Stage 2 & \textbf{98.60} & \textbf{99.60} & \textbf{98.20} & \textbf{98.60} & \textbf{98.75} \\
\midrule
\multicolumn{6}{@{}l}{\textbf{QwenVLA}} \\
Initialization & 89.40 & 94.40 & 88.40 & 52.00 & 81.05 \\
Teachers & \textbf{99.20} & \underline{99.40} & \underline{97.60} & 90.80 & \underline{96.75} \\
\shortstack[l]{SFT \citep{community2026starvla}} & \underline{97.80} & 98.80 & 97.40 & \underline{92.00} & 96.50 \\
Stage 1 & 97.20 & \underline{99.40} & 93.40 & 79.00 & 92.25 \\
Stage 1 + Stage 2 & \textbf{99.20} & \textbf{100.00} & \textbf{98.60} & \textbf{93.40} & \textbf{97.80} \\
\midrule
\multicolumn{6}{@{}l}{{\boldmath$\pi_{0.5}$}} \\
Initialization & 78.20 & 92.80 & 88.60 & 28.40 & 72.00 \\
Teachers & 97.80 & \textbf{100.00} & 93.80 & 91.40 & 95.75 \\
\shortstack[l]{SFT \citep{openpi_libero}} & \textbf{98.80} & 98.20 & \underline{98.00} & \underline{92.40} & \underline{96.85} \\
Stage 1 & 94.20 & 98.00 & 94.60 & 83.40 & 92.55 \\
Stage 1 + Stage 2 & \underline{98.61} & \underline{99.90} & \textbf{98.60} & \textbf{94.44} & \textbf{97.89} \\
\bottomrule
\end{tabular*}
\end{table}

\begin{table}[htbp]
\centering
\caption{Success rates (\%) on RoboCasa.}
\label{tab:main-robocasa}
\fontsize{8}{9.2}\selectfont
\setlength{\tabcolsep}{2pt}
\renewcommand{\arraystretch}{1.0}
\setlength{\aboverulesep}{1.2pt}
\setlength{\belowrulesep}{1.2pt}
\begin{tabular*}{\columnwidth}{@{\extracolsep{\fill}}>{\raggedright\arraybackslash}p{0.39\columnwidth}
    >{\raggedleft\arraybackslash}p{0.13\columnwidth}
    >{\raggedleft\arraybackslash}p{0.13\columnwidth}
    >{\raggedleft\arraybackslash}p{0.13\columnwidth}
    >{\raggedleft\arraybackslash}p{0.08\columnwidth}
    @{}}
\toprule
\noalign{\vskip 2.25pt}
\multicolumn{5}{@{}l}{\textbf{RoboCasa Kitchen}} \\
\textbf{GR00T N1.5} & PnP & Articulated & \shortstack[r]{Appliance--\\Coffee} & \textbf{Avg.} \\
\midrule
Initialization & 41.75 & 63.25 & 67.75 & 57.58 \\
Teachers & 50.50 & 68.75 & 68.75 & 62.67 \\
\shortstack[l]{SFT \citep{koo2026hamlet}} & \underline{53.75} & \underline{73.25} & 65.25 & \underline{64.08} \\
Stage 1 & 49.25 & 62.75 & \underline{74.75} & 62.25 \\
Stage 1 + Stage 2 & \textbf{63.00} & \textbf{75.25} & \textbf{88.25} & \textbf{75.50} \\
\midrule
\noalign{\vskip 2.25pt}
\multicolumn{5}{@{}l}{\textbf{RoboCasa GR1}} \\
\textbf{QwenVLA} & Articulated & Tableware & Storage & \textbf{Avg.} \\
\midrule
Initialization & 32.00 & 22.00 & 16.67 & 22.50 \\
Teachers & \textbf{52.00} & 53.78 & \underline{42.44} & \underline{49.08} \\
\shortstack[l]{SFT \citep{community2026starvla}} & \underline{50.33} & \textbf{55.78} & 38.22 & 47.83 \\
Stage 1 & \underline{50.33} & 52.44 & 41.33 & 47.75 \\
Stage 1 + Stage 2 & 45.33 & \underline{54.44} & \textbf{48.22} & \textbf{49.83} \\
\bottomrule
\end{tabular*}
\end{table}

\begin{figure}[t]
    \centering
    \includegraphics[width=\columnwidth]{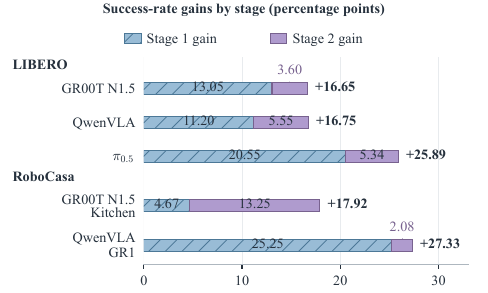}
    \caption{Success-rate gains from each training stage,
    measured in percentage points. Bar segments show the
    individual stage gains; end labels show the total gain
    over initialization.}
    \label{fig:stage-gains}
\end{figure}

Tables~\ref{tab:main-libero} and~\ref{tab:main-robocasa} show that Stage 1 improves mean success rates in all five settings. These improvements support the effectiveness of multi-teacher distillation in consolidating task expertise into a shared model. Stage 2 builds on these capabilities to further improve task performance, and the full framework exceeds the teachers' aggregate success rates in all five settings. Figure~\ref{fig:stage-gains} further highlights how the gains from the two stages vary across models and environments. Even on LIBERO, where Stage 1 already achieves high success rates, joint optimization in Stage 2 provides further gains, indicating that capability consolidation leaves room for additional improvements in task success. Together, these results support combining expert capability consolidation with subsequent joint optimization to improve multitask performance within a shared model.

\subsection{Ablation Studies}
\label{sec:ablations}

\begin{figure}[t]
    \centering
    \includegraphics[width=\columnwidth]{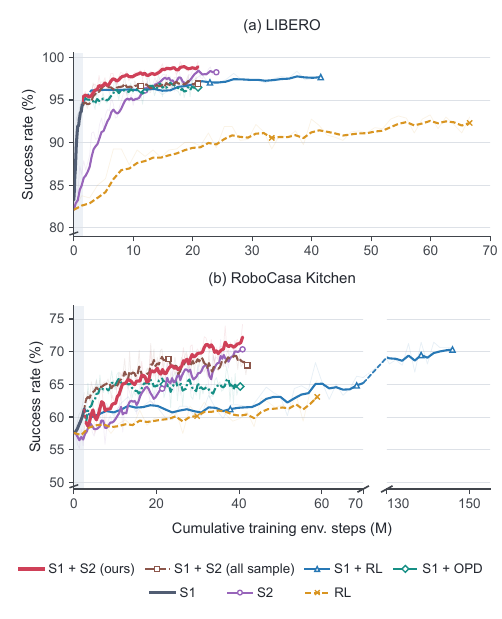}
    \caption{
    Ablation studies with GR00T N1.5. Thick curves show exponential moving averages (coefficient 0.3); faint curves show raw evaluations. Shaded regions mark Stage 1 for the two-stage variants. The RoboCasa Kitchen horizontal axis omits 70--130 million steps.
    }
    \label{fig:ablations}
\end{figure}

We conduct ablations with GR00T N1.5 on LIBERO and RoboCasa Kitchen to examine the two training stages and failure-gated distillation in Stage 2. We compare the full two-stage framework (S1 + S2, ours) with direct RL, training with Stage 2 only (S2), and Stage 1 followed by either RL alone (S1 + RL) or teacher distillation alone (S1 + OPD). We also include an all-sample variant that applies teacher distillation to both successful and failed trajectories during Stage 2 (S1 + S2, all sample). Figure~\ref{fig:ablations} presents the learning curves. The horizontal axis measures cumulative training environment steps in millions, including interactions from both stages for the two-stage variants.

Stage 1 rapidly improves task success through dense teacher supervision, providing a useful initialization with limited environment interaction. On both benchmarks, the two-stage framework reaches high success rates earlier than applying the Stage 2 objective directly to the SFT checkpoint. S1 + RL also shows better learning progress than direct RL, indicating that prior capability consolidation benefits subsequent optimization using environmental rewards alone. These results highlight the advantage of distilling teacher
capabilities before subsequent optimization.

Starting from the same Stage 1 checkpoint, teacher distillation alone plateaus relatively early, whereas RL alone improves more slowly. Both objectives yield gains individually, but their combination achieves higher success rates at comparable interaction budgets. At approximately 20 million environment steps on LIBERO, the smoothed success rate is 98.7\% for our framework, compared with 96.4\% for S1 + OPD and 96.9\% for S1 + RL. On RoboCasa Kitchen, the full framework reaches  approximately 71\% at 40 million steps, versus approximately 65\% for S1 + OPD and 61\% for S1 + RL. These results favor retaining both learning signals after teacher distillation.

The benefits of failure-gated distillation in Stage 2 are clearest later in training. On RoboCasa Kitchen, the all-sample variant learns faster initially, but its success rate subsequently plateaus. Our framework continues to improve
and achieves higher success rates later in training. On LIBERO, the all-sample variant also plateaus earlier, while failure-gated distillation yields further gains. These results highlight the advantage of focusing teacher guidance on failed trajectories, preserving dense supervision while reducing direct constraints on successful behaviors.

\begin{figure}[htbp]
    \centering
    \includegraphics[width=\columnwidth]
    {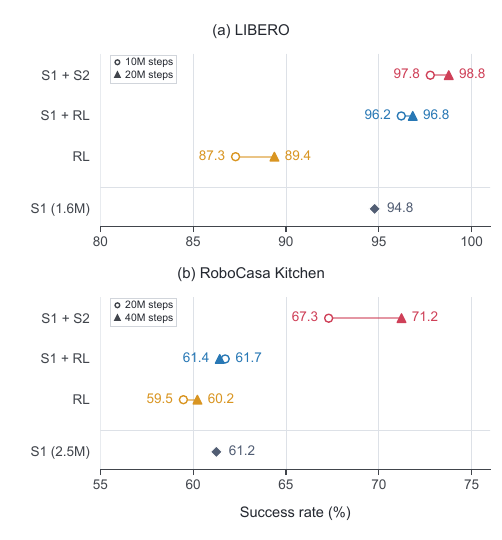}
    \caption{
    Sample efficiency measured by success rates at fixed numbers
    of training environment steps (M: million).
    Values are taken from smoothed learning curves.
    Two-stage totals include Stage 1.
    Diamonds mark Stage 1 endpoints at their actual step counts.
    }
    \label{fig:fixed-budget-success}
\end{figure}

\subsection{Sample Efficiency and Hyperparameter Sensitivity}
\label{sec:efficiency-sensitivity}

We further analyze the sample efficiency of the framework and its sensitivity to the Stage 2 distillation weight $\beta$. All experiments in this subsection use GR00T N1.5. 

Figure~\ref{fig:fixed-budget-success} reports success rates of different training methods at fixed interaction budgets from smoothed learning curves, with all Stage 1 interactions included in the budgets of two-stage methods. For reference, the figure also shows Stage 1 endpoints at their actual training budgets. Stage 1 reaches 94.8\% on LIBERO after approximately 1.6 million environment steps and 61.2\% on RoboCasa Kitchen after 2.5 million steps, illustrating efficient early learning from dense teacher supervision.

The full framework outperforms direct RL at every budget
considered. Stage 1 already achieves substantial gains with
limited environment interaction. On LIBERO, it reaches a
smoothed success rate of 94.8\% after approximately 1.6 million
steps, exceeding the 89.4\% attained by direct RL after
20 million steps. On RoboCasa Kitchen, Stage 1 reaches
61.2\% after approximately 2.5 million steps, compared with
60.2\% for direct RL after 40 million steps. These results
illustrate how dense teacher supervision establishes a
strong initialization with relatively few environment
interactions.

Building on this initialization, the full framework reaches
97.8\% on LIBERO with 10 million environment steps and
67.3\% on RoboCasa Kitchen with 20 million steps.
In both cases, it achieves higher success rates than
direct RL using half as many environment steps.
At matched budgets, it also outperforms S1 + RL by
approximately 2.0 percentage points on LIBERO at
20 million steps and 9.8 percentage points on RoboCasa
Kitchen at 40 million steps. These comparisons indicate
that both initial teacher distillation and continued
teacher guidance in Stage 2 contribute to improved
sample efficiency.

\begin{figure}[tbp]
    \centering
    \includegraphics[width=\columnwidth]
    {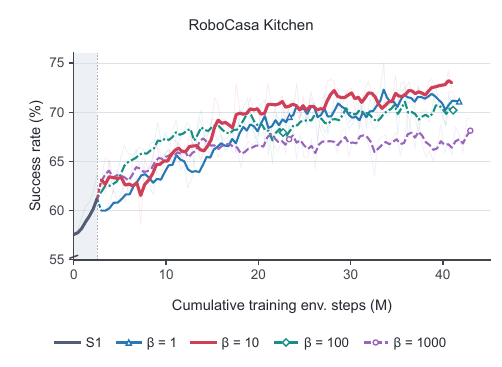}
    \caption{
    Sensitivity to the Stage 2 distillation weight $\beta$
    on RoboCasa Kitchen.
    The shaded region denotes Stage 1.
    Thick and faint lines show smoothed and unsmoothed
    success rates.
    }
    \label{fig:beta-sensitivity}
\end{figure}

We further examine sensitivity to the Stage 2 distillation weight $\beta$ on RoboCasa Kitchen, comparing $\beta\in\{1,10,100,1000\}$. As shown in Figure~\ref{fig:beta-sensitivity},
$\beta=100$ achieves higher success rates early in training,
but $\beta=10$ performs better later.
With $\beta=1000$, performance plateaus relatively early
and improves little thereafter. Thus, the strength of teacher guidance affects joint optimization, and stronger supervision does not necessarily yield better final performance.
In this setting, $\beta=10$ provides a favorable balance between teacher guidance and optimization using environmental rewards.

\section{Conclusion}
\label{sec:conclusion}

We propose a two-stage framework for post training flow-based embodied models via multi-teacher on policy distillation. The distillation objectives in both stages follow the velocity regression formulation of native conditional flow matching. Stage 1 uses MOPD to efficiently consolidate expert capabilities.
Stage 2 combines RL with teacher distillation adapted to the transition and noise scales of stochastic flow sampling. Teacher guidance is restricted to failed trajectories, supplementing sparse rewards with dense supervision without directly constraining successful behaviors.
Experiments across three embodied models and benchmarks show that the framework improves mean success rates across tasks and surpasses the aggregate
performance of the corresponding teachers. Ablations highlight the benefits of the two-stage design and failure-gated distillation.
Sample efficiency comparisons further show that the framework
achieves higher success rates than direct RL with fewer environment interactions.



\bibliographystyle{icml2027}
\bibliography{references}

@article{mimo,
  author  = {{LLM-Core Xiaomi}},
  title   = {{MiMo-V2-Flash} Technical Report},
  journal = {arXiv preprint arXiv:2601.02780},
  year    = {2026}
}

@article{deepseek,
  author  = {{DeepSeek-AI}},
  title   = {{DeepSeek-V4}: Towards Highly Efficient Million-Token Context Intelligence},
  journal = {arXiv preprint arXiv:2606.19348},
  year    = {2026}
}

@article{kimi,
  author  = {{Kimi Team}},
  title   = {{Kimi K3}: Open Frontier Intelligence},
  journal = {arXiv preprint arXiv:2607.24653},
  year    = {2026}
}

@article{qwen3,
  author  = {{Qwen Team}},
  title   = {{Qwen3} Technical Report},
  journal = {arXiv preprint arXiv:2505.09388},
  year    = {2025}
}

@article{vlaopd,
  author  = {Zhong, Zhide and Yan, Haodong and Li, Junfeng
             and He, Junjie and Zhang, Tianran and Li, Haoang},
  title   = {{VLA-OPD}: Bridging Offline {SFT} and Online {RL}
             for Vision-Language-Action Models via On-Policy Distillation},
  journal = {arXiv preprint arXiv:2603.26666},
  year    = {2026}
}

@article{roadmopd,
  author  = {Tian, Shi-Yu and Liu, Yong and Liu, Ximan
             and Zhu, Tianrui and Bai, Sule and Yang, Xinwei
             and Liu, Ruiqi and Yu, Kun-Yang and Yang, Ming
             and Zhou, Zhi and Li, Yu-Feng and Sun, Yipeng
             and Gu, Simiu},
  title   = {Beyond Domain-Routed Supervision: Cross-Teacher Guarding
             for Multi-Teacher On-Policy Distillation},
  journal = {arXiv preprint arXiv:2610.13521},
  year    = {2026}
}

@article{pmopd,
  author  = {Liu, Youzhi and Zheng, Ruobing and Tong, Boyuan
             and Li, Tianqi and Li, Pingqi and Bi, Hanbo
             and Yuan, Yi and Chen, Jingdong},
  title   = {{PMOPD}: Task Ordering, Cycling, and Parameter-Update
             Subspace Protection in Multi-Teacher On-Policy Distillation},
  journal = {arXiv preprint arXiv:2609.34605},
  year    = {2026}
}

@article{flowmopd,
  author  = {Ni, Lichao and Liu, Yuhan and Li, Pengfei
             and Fang, Guanghuan},
  title   = {Multi-Teacher On-Policy Distillation for Flow-Based
             Vision-Language-Action Models},
  journal = {arXiv preprint arXiv:2610.21522},
  year    = {2026}
}

@article{vlamopd,
  author  = {Lv, Yongxuan and Fang, Zhen and Man, Lichen
             and Li, Hao and Li, Pu and Wang, Hengyi
             and Cao, Langfeng and Chen, Lin and Huang, Jie
             and Su, Xiu and You, Shan and Wang, Fei
             and Zhao, Feng},
  title   = {{VLA-MOPD}: On-Policy Distillation for Multi-Task
             {VLA} Policy Consolidation},
  journal = {arXiv preprint arXiv:2611.12286},
  year    = {2026}
}

@inproceedings{pi0,
  author = {Black, Kevin and Brown, Noah and Driess, Danny and
            Esmail, Adnan and Equi, Michael Robert and Finn, Chelsea and
            Fusai, Niccolo and Groom, Lachy and Hausman, Karol and
            Ichter, Brian and Jakubczak, Szymon and Jones, Tim and
            Ke, Liyiming and Levine, Sergey and Li-Bell, Adrian and
            Mothukuri, Mohith and Nair, Suraj and Pertsch, Karl and
            Shi, Lucy Xiaoyang and Smith, Laura and Tanner, James and
            Vuong, Quan and Walling, Anna and Wang, Haohuan and
            Zhilinsky, Ury},
  title = {{$\pi_0$}: A Vision-Language-Action Flow Model for General Robot Control},
  booktitle = {Proceedings of Robotics: Science and Systems},
  year = {2025}
}

@article{dreamzero,
  author = {Ye, Seonghyeon and Ge, Yunhao and Zheng, Kaiyuan and
            Gao, Shenyuan and Yu, Sihyun and Kurian, George and
            Indupuru, Suneel and Tan, You Liang and Zhu, Chuning and
            Xiang, Jiannan and Malik, Ayaan and Lee, Kyungmin and
            Liang, William and Ranawaka, Nadun and Gu, Jiasheng and
            Xu, Yinzhen and Wang, Guanzhi and Hu, Fengyuan and
            Narayan, Avnish and Bjorck, Johan and Wang, Jing and
            Kim, Gwanghyun and Niu, Dantong and Zheng, Ruijie and
            Xie, Yuqi and Wu, Jimmy and Wang, Qi and Julian, Ryan and
            Xu, Danfei and Du, Yilun and Chebotar, Yevgen and
            Reed, Scott and Kautz, Jan and Zhu, Yuke and
            Fan, Linxi {``Jim''} and Jang, Joel},
  title = {World Action Models are Zero-shot Policies},
  journal = {arXiv preprint arXiv:2602.15922},
  year = {2026}
}

@inproceedings{nestedvla,
  author = {Du, Guodong and Lin, Wanyu},
  title = {{NestedVLA}: Learning to Consolidate and Generate Skills for
           Vision-Language-Action Models},
  booktitle = {Advances in Neural Information Processing Systems},
  year = {2026}
}

@article{robion,
  author = {Adamopoulos, Dionysios and Chanpaisit, Nattapol and
            Fakhri, Basel and Giannoula, Christina},
  title = {Efficient Vision-Language-Action Management and Serving
           for Robot Factories},
  journal = {arXiv preprint arXiv:2609.12075},
  year = {2026}
}

@inproceedings{dagger,
  author = {Ross, Stephane and Gordon, Geoffrey and Bagnell, Drew},
  title = {A Reduction of Imitation Learning and Structured Prediction
           to No-Regret Online Learning},
  booktitle = {Proceedings of the Fourteenth International Conference
               on Artificial Intelligence and Statistics},
  series = {Proceedings of Machine Learning Research},
  volume = {15},
  pages = {627--635},
  year = {2011}
}

@inproceedings{vlarl,
  author = {Lu, Guanxing and Guo, Wenkai and Zhang, Chubin and
            Zhou, Yuheng and Jiang, Haonan and Gao, Zifeng and
            Tang, Yansong and Wang, Ziwei},
  title = {{VLA-RL}: Towards Masterful and General Robotic
           Manipulation with Scalable Reinforcement Learning},
  booktitle = {IEEE/RSJ International Conference on Intelligent
               Robots and Systems},
  year = {2026}
}

@inproceedings{mtopt,
  author = {Kalashnikov, Dmitry and Varley, Jake and
            Chebotar, Yevgen and Swanson, Benjamin and
            Jonschkowski, Rico and Finn, Chelsea and
            Levine, Sergey and Hausman, Karol},
  title = {Scaling Up Multi-Task Robotic Reinforcement Learning},
  booktitle = {Proceedings of the 5th Conference on Robot Learning},
  series = {Proceedings of Machine Learning Research},
  volume = {164},
  pages = {557--575},
  year = {2022}
}

@inproceedings{droid,
  author = {Alexander Khazatsky and Karl Pertsch and Suraj Nair and
            Ashwin Balakrishna and Sudeep Dasari and
            Siddharth Karamcheti and Soroush Nasiriany and
            Mohan Kumar Srirama and Lawrence Yunliang Chen and
            Kirsty Ellis and Peter David Fagan and Joey Hejna and
            Masha Itkina and Marion Lepert and Yecheng Jason Ma and
            Patrick Tree Miller and Jimmy Wu and Suneel Belkhale and
            Shivin Dass and Huy Ha and Arhan Jain and Abraham Lee and
            Youngwoon Lee and Marius Memmel and Sungjae Park and
            Ilija Radosavovic and Kaiyuan Wang and Albert Zhan and
            Kevin Black and Cheng Chi and Kyle Beltran Hatch and
            Shan Lin and Jingpei Lu and Jean Mercat and
            Abdul Rehman and Pannag R. Sanketi and Archit Sharma and
            Cody Simpson and Quan Vuong and Homer Rich Walke and
            Blake Wulfe and Ted Xiao and Jonathan Heewon Yang and
            Arefeh Yavary and Tony Z. Zhao and Christopher Agia and
            Rohan Baijal and Mateo Guaman Castro and Daphne Chen and
            Qiuyu Chen and Trinity Chung and Jaimyn Drake and
            Ethan Paul Foster and Jensen Gao and
            David Antonio Herrera and Minho Heo and Kyle Hsu and
            Jiaheng Hu and Donovon Jackson and Charlotte Le and
            Yunshuang Li and Roy Lin and Zehan Ma and
            Abhiram Maddukuri and Suvir Mirchandani and Daniel Morton and
            Tony Nguyen and Abigail O'Neill and Rosario Scalise and
            Derick Seale and Victor Son and Stephen Tian and Emi Tran and
            Andrew E. Wang and Yilin Wu and Annie Xie and Jingyun Yang and
            Patrick Yin and Yunchu Zhang and Osbert Bastani and
            Glen Berseth and Jeannette Bohg and Ken Goldberg and
            Abhinav Gupta and Abhishek Gupta and Dinesh Jayaraman and
            Joseph J. Lim and Jitendra Malik and
            Roberto Mart{\'i}n-Mart{\'i}n and Subramanian Ramamoorthy and
            Dorsa Sadigh and Shuran Song and Jiajun Wu and
            Michael C. Yip and Yuke Zhu and Thomas Kollar and
            Sergey Levine and Chelsea Finn},
  title = {{DROID}: A Large-Scale In-The-Wild Robot Manipulation Dataset},
  booktitle = {Proceedings of Robotics: Science and Systems},
  year = {2024}
}

@inproceedings{flowmatching,
  author = {Lipman, Yaron and Chen, Ricky T. Q. and
            Ben-Hamu, Heli and Nickel, Maximilian and Le, Matt},
  title = {Flow Matching for Generative Modeling},
  booktitle = {The Eleventh International Conference on
               Learning Representations},
  year = {2023}
}

@inproceedings{gkd,
  author = {Agarwal, Rishabh and Vieillard, Nino and
            Zhou, Yongchao and Stanczyk, Piotr and
            Ramos Garea, Sabela and Geist, Matthieu and
            Bachem, Olivier},
  title = {On-Policy Distillation of Language Models:
           Learning from Self-Generated Mistakes},
  booktitle = {The Twelfth International Conference on
               Learning Representations},
  year = {2024}
}

@inproceedings{minillm,
  author = {Gu, Yuxian and Dong, Li and Wei, Furu and Huang, Minlie},
  title = {{MiniLLM}: Knowledge Distillation of Large Language Models},
  booktitle = {The Twelfth International Conference on
               Learning Representations},
  year = {2024}
}

@article{glm5,
  author = {{GLM-5 Team}},
  title = {{GLM-5}: from Vibe Coding to Agentic Engineering},
  journal = {arXiv preprint arXiv:2602.15763},
  year = {2026}
}

@article{nemotroncascade2,
  author = {Yang, Zhuolin and Liu, Zihan and Chen, Yang and
            Dai, Wenliang and Wang, Boxin and Lin, Sheng-Chieh and
            Lee, Chankyu and Chen, Yangyi and Jiang, Dongfu and
            He, Jiafan and Pi, Renjie and Lam, Grace and
            Lee, Nayeon and Bukharin, Alexander and
            Shoeybi, Mohammad and Catanzaro, Bryan and Ping, Wei},
  title = {{Nemotron-Cascade 2}: Post-Training {LLMs} with
           Cascade {RL} and Multi-Domain On-Policy Distillation},
  journal = {arXiv preprint arXiv:2603.19220},
  year = {2026}
}

@inproceedings{irevla,
  author    = {Guo, Yanjiang and Zhang, Jianke and Chen, Xiaoyu
               and Ji, Xiang and Wang, Yen-Jen and Hu, Yucheng
               and Chen, Jianyu},
  title     = {Improving Vision-Language-Action Model with Online
               Reinforcement Learning},
  booktitle = {IEEE International Conference on Robotics and Automation},
  year      = {2025}
}

@inproceedings{simplevlarl,
  author    = {Li, Haozhan and Zuo, Yuxin and Yu, Jiale
               and Zhang, Yuhao and Yang, Zhaohui and Zhang, Kaiyan
               and Zhu, Xuekai and Zhang, Yuchen and Chen, Tianxing
               and Cui, Ganqu and Wang, Dehui and Luo, Dingxiang
               and Fan, Yuchen and Sun, Youbang and Zeng, Jia
               and Pang, Jiangmiao and Zhang, Shanghang and Wang, Yu
               and Mu, Yao and Zhou, Bowen and Ding, Ning},
  title     = {{SimpleVLA-RL}: Scaling {VLA} Training via
               Reinforcement Learning},
  booktitle = {The Fourteenth International Conference on
               Learning Representations},
  year      = {2026}
}

@inproceedings{reinflow,
  author    = {Zhang, Tonghe and Yu, Chao and Su, Sichang and Wang, Yu},
  title     = {{ReinFlow}: Fine-tuning Flow Matching Policy with
               Online Reinforcement Learning},
  booktitle = {Advances in Neural Information Processing Systems},
  volume    = {38},
  year      = {2025}
}

@article{pirl,
  author  = {Chen, Kang and Liu, Zhihao and Zhang, Tonghe
             and Guo, Zhen and Xu, Si and Lin, Hao
             and Zang, Hongzhi and Li, Xiang and Wei, Bingwen
             and Zhou, Jiakai and Zhang, Quanlu and Yu, Zhaofei
             and Fan, Guoliang and Huang, Tiejun and Wang, Yu
             and Yu, Chao},
  title   = {{$\pi_{\mathrm{RL}}$}: Online {RL} Fine-tuning for
             Flow-based Vision-Language-Action Models},
  journal = {arXiv preprint arXiv:2510.25889},
  year    = {2026}
}

@inproceedings{flowmatchingpg,
  author    = {McAllister, David and Ge, Songwei and Yi, Brent
               and Kim, Chung Min and Weber, Ethan and Choi, Hongsuk
               and Feng, Haiwen and Kanazawa, Angjoo},
  title     = {Flow Matching Policy Gradients},
  booktitle = {The Fourteenth International Conference on
               Learning Representations},
  year      = {2026}
}

@inproceedings{lyufpo,
  author    = {Lyu, Mingyang and Sun, Yinqian and Lin, Erliang
               and Li, Huangrui and Chen, Ruolin
               and Zhao, Feifei and Zeng, Yi},
  title     = {Reinforcement Fine-Tuning of Flow-Matching Policies
               for Vision-Language-Action Models},
  booktitle = {IEEE International Conference on Robotics and Automation},
  year      = {2026}
}

@article{qvgm,
  author  = {Wang, Ziqian and Zhang, Rui and Liu, Yitian
             and Mao, Xingjian and Wang, Minqian and Mu, Yao},
  title   = {{Q-VGM}: Q-Guided Value-Gradient Matching for
             Offline-to-Online {RL} of Flow-Matching {VLA} Policies},
  journal = {arXiv preprint arXiv:2606.08015},
  year    = {2026}
}

@inproceedings{dsrl,
  author    = {Wagenmaker, Andrew and Zhang, Yunchu
               and Nakamoto, Mitsuhiko and Park, Seohong
               and Yagoub, Waleed and Nagabandi, Anusha
               and Gupta, Abhishek and Levine, Sergey},
  title     = {Steering Your Diffusion Policy with Latent Space
               Reinforcement Learning},
  booktitle = {Proceedings of The 9th Conference on Robot Learning},
  series    = {Proceedings of Machine Learning Research},
  volume    = {305},
  pages     = {258--282},
  publisher = {PMLR},
  year      = {2025}
}

@inproceedings{diffusionopd,
  author    = {Li, Quanhao and Yu, Junqiu and Jiang, Kaixun
               and Wei, Yujie and Xing, Zhen and Li, Pandeng
               and Chu, Ruihang and Zhang, Shiwei and Liu, Yu
               and Wu, Zuxuan},
  title     = {{DiffusionOPD}: A Unified Perspective of
               On-Policy Distillation in Diffusion Models},
  booktitle = {SIGGRAPH Asia 2026 Conference Papers},
  year      = {2026}
}

@article{danceopd,
  author  = {Zhou, Wei and Zhu, Xiongwei and Xu, Zelin
             and Dong, Bo and Gong, Lixue and Liang, Yongyuan
             and Chu, Meng and Qu, Leigang and Kong, Lingdong
             and Liu, Wei and Chua, Tat-Seng},
  title   = {{DanceOPD}: On-Policy Generative Field Distillation},
  journal = {arXiv preprint arXiv:2606.27377},
  year    = {2026}
}

@article{flowcts,
  author  = {Ye, Kaiyang and Ge, Yuan and Zhang, Junxiang
             and Li, Bei and Zhu, Ziming and Zhao, Haishu
             and Liu, Xiaoqian and Wang, Chenglong
             and Zhu, Jingbo and Yu, Zhengtao and Xiao, Tong},
  title   = {{FlowCTS}: On-policy Continuous Trajectory
             Supervision of Flow Models},
  journal = {arXiv preprint arXiv:2607.24522},
  year    = {2026}
}

@article{repair,
  author  = {Jiang, Zhennan and Xie, Tingying and Ji, Shilong
             and Yu, Shu'ang and Zhou, Tianxing and Chen, Yuhui
             and Gao, Jiahuan and Xie, Kaizhi and Xu, Yuanfan
             and Gao, Feng and Wang, Yu and Li, Haoran
             and Zhao, Dongbin and Yu, Chao},
  title   = {Beyond Policy Patches: Distilling {RL} Specialists
             into Vision-Language-Action Policies},
  journal = {arXiv preprint arXiv:2611.02777},
  year    = {2026}
}

@inproceedings{flowgrpo,
  author    = {Liu, Jie and Liu, Gongye and Liang, Jiajun
               and Li, Yangguang and Liu, Jiaheng
               and Wang, Xintao and Wan, Pengfei
               and Zhang, Di and Ouyang, Wanli},
  title     = {{Flow-GRPO}: Training Flow Matching Models
               via Online {RL}},
  booktitle = {Advances in Neural Information Processing Systems},
  volume    = {38},
  year      = {2025}
}

@article{grootn1,
  title = {{GR00T N1}: An Open Foundation Model for Generalist Humanoid Robots},
  author = {{NVIDIA} and Johan Bjorck and Fernando Casta{\~n}eda
    and Nikita Cherniadev and Xingye Da and Runyu Ding
    and Linxi "Jim" Fan and Yu Fang and Dieter Fox and Fengyuan Hu
    and Spencer Huang and Joel Jang and Zhenyu Jiang and Jan Kautz
    and Kaushil Kundalia and Lawrence Lao and Zhiqi Li and Zongyu Lin
    and Kevin Lin and Guilin Liu and Edith Llontop and Loic Magne
    and Ajay Mandlekar and Avnish Narayan and Soroush Nasiriany
    and Scott Reed and You Liang Tan and Guanzhi Wang and Zu Wang
    and Jing Wang and Qi Wang and Jiannan Xiang and Yuqi Xie
    and Yinzhen Xu and Zhenjia Xu and Seonghyeon Ye and Zhiding Yu
    and Ao Zhang and Hao Zhang and Yizhou Zhao and Ruijie Zheng
    and Yuke Zhu},
  journal = {arXiv preprint arXiv:2503.14734},
  year = {2025}
}

@article{qwenvla,
  title = {{Qwen-VLA}: Unifying Vision-Language-Action Modeling across Tasks, Environments, and Robot Embodiments},
  author = {Qiuyue Wang and Mingsheng Li and Jian Guan and Jinhui Ye
    and Sicheng Xie and Yitao Liu and Junhao Chen and Zhixuan Liang
    and Jie Zhang and Xintong Hu and Xuhong Huang and Pei Lin
    and Junyang Lin and Dayiheng Liu and Shuai Bai and Jingren Zhou
    and Jiazhao Zhang and Haoqi Yuan and Gengze Zhou and Hang Yin
    and Ye Wang and Yiyang Huang and Zixing Lei and Wujian Peng
    and Delin Chen and Yingming Zheng and Jingyang Fan
    and Xianwei Zhuang and Xin Zhou and Haoyang Li and Anzhe Chen
    and Tong Zhang and Xuejing Liu and Yuchong Sun and Ruizhe Chen
    and Zhaohai Li and Chenxu L{\"u} and Zhibo Yang and Tao Yu
    and Xionghui Chen},
  journal = {arXiv preprint arXiv:2605.30280},
  year = {2026}
}

@inproceedings{pi05,
  title = {{$\pi_{0.5}$}: a Vision-Language-Action Model with Open-World Generalization},
  author = {Black, Kevin and Brown, Noah and Darpinian, James
    and Dhabalia, Karan and Driess, Danny and Esmail, Adnan
    and Equi, Michael Robert and Finn, Chelsea and Fusai, Niccolo
    and Galliker, Manuel Y. and Ghosh, Dibya and Groom, Lachy
    and Hausman, Karol and ichter, brian and Jakubczak, Szymon
    and Jones, Tim and Ke, Liyiming and LeBlanc, Devin
    and Levine, Sergey and Li-Bell, Adrian and Mothukuri, Mohith
    and Nair, Suraj and Pertsch, Karl and Ren, Allen Z.
    and Shi, Lucy Xiaoyang and Smith, Laura
    and Springenberg, Jost Tobias and Stachowicz, Kyle
    and Tanner, James and Vuong, Quan and Walke, Homer
    and Walling, Anna and Wang, Haohuan and Yu, Lili
    and Zhilinsky, Ury},
  booktitle = {Proceedings of the 9th Conference on Robot Learning},
  series = {Proceedings of Machine Learning Research},
  volume = {305},
  pages = {17--40},
  publisher = {PMLR},
  year = {2025}
}

@inproceedings{libero,
  title = {{LIBERO}: Benchmarking Knowledge Transfer for Lifelong Robot Learning},
  author = {Liu, Bo and Zhu, Yifeng and Gao, Chongkai and Feng, Yihao
    and Liu, Qiang and Zhu, Yuke and Stone, Peter},
  booktitle = {Advances in Neural Information Processing Systems},
  volume = {36},
  pages = {44776--44791},
  year = {2023}
}

@inproceedings{robocasa,
  title = {{RoboCasa}: Large-Scale Simulation of Household Tasks for Generalist Robots},
  author = {Soroush Nasiriany and Abhiram Maddukuri and Lance Zhang
    and Adeet Parikh and Aaron Lo and Abhishek Joshi
    and Ajay Mandlekar and Yuke Zhu},
  booktitle = {Proceedings of Robotics: Science and Systems},
  year = {2024}
}

@article{dial,
  title = {{DIAL}: Decoupling Intent and Action via Latent
           World Modeling for End-to-End {VLA}},
  author = {Chen, Yi and Ge, Yuying and Zhou, Hui
            and Ding, Mingyu and Ge, Yixiao and Liu, Xihui},
  journal = {arXiv preprint arXiv:2603.29844},
  year = {2026}
}

@inproceedings{koo2026hamlet,
  author = {Koo, Myungkyu and Choi, Daewon and Kim, Taeyoung and
            Lee, Kyungmin and Kim, Changyeon and Seo, Younggyo and
            Shin, Jinwoo},
  title = {{HAMLET}: Switch Your Vision-Language-Action Model
           into a History-Aware Policy},
  booktitle = {The Fourteenth International Conference on
               Learning Representations},
  year = {2026}
}

@article{community2026starvla,
  author = {{StarVLA Community}},
  title = {{StarVLA}: A Lego-like Codebase for
           Vision-Language-Action Model Developing},
  journal = {arXiv preprint arXiv:2604.05014},
  year = {2026}
}

@misc{openpi_libero,
  author = {{Physical Intelligence}},
  title = {{openpi}: {LIBERO} Benchmark},
  howpublished = {GitHub repository documentation},
  year = {2025}
}

\clearpage
\appendix
\onecolumn
\section{Implementation Details}
\label{app:implementation}

\renewcommand{\thetable}{\thesection.\arabic{table}}
\setcounter{table}{0}
This appendix provides the task groups for RoboCasa Kitchen and RoboCasa GR1, and the two-stage training configurations for LIBERO and RoboCasa.

\subsection{Task Groups}
\label{app:task-groups}

We group tasks in each RoboCasa benchmark by similarity and train one expert teacher per group. We train a single shared model for each benchmark and select the teacher according to the task group.  Tables~\ref{tab:app-kitchen-tasks} and~\ref{tab:app-gr1-tasks}
list the task assignments.

\begingroup
\small
\setlength{\tabcolsep}{4pt}
\renewcommand{\arraystretch}{1.25}
\begin{longtable}{@{}p{0.20\linewidth}>{\raggedright\arraybackslash}p{0.10\linewidth}p{\dimexpr0.74\linewidth-4\tabcolsep\relax}@{}}
\caption{Task groups in RoboCasa Kitchen.}\label{tab:app-kitchen-tasks}\\
\toprule
\textbf{Task group} & \textbf{No. of Tasks} & \textbf{Task names} \\
\midrule
\endfirsthead
\multicolumn{3}{l}{\tablename~\thetable\ (continued)}\\
\toprule
\textbf{Task group} & \textbf{Tasks} & \textbf{Task names} \\
\midrule
\endhead
\midrule
\multicolumn{3}{r}{Continued on next page}\\
\endfoot
\bottomrule
\endlastfoot
PnP & 8 & \begin{tabular}[t]{@{}>{\ttfamily}l@{}}
PnPCabToCounter \\
PnPCounterToCab \\
PnPCounterToMicrowave \\
PnPCounterToSink \\
PnPCounterToStove \\
PnPMicrowaveToCounter \\
PnPSinkToCounter \\
PnPStoveToCounter
\end{tabular} \\
\midrule
Articulated & 8 & \begin{tabular}[t]{@{}>{\ttfamily}l@{}}
CloseDoubleDoor \\
CloseDrawer \\
CloseSingleDoor \\
OpenDoubleDoor \\
OpenDrawer \\
OpenSingleDoor \\
TurnOffStove \\
TurnOnStove
\end{tabular} \\
\midrule
Appliance--Coffee & 8 & \begin{tabular}[t]{@{}>{\ttfamily}l@{}}
CoffeePressButton \\
CoffeeServeMug \\
CoffeeSetupMug \\
TurnOffMicrowave \\
TurnOffSinkFaucet \\
TurnOnMicrowave \\
TurnOnSinkFaucet \\
TurnSinkSpout
\end{tabular} \\
\end{longtable}
\endgroup

\clearpage
For readability, the GR1 task names in
Table~\ref{tab:app-gr1-tasks} omit the common prefix
\texttt{gr1\_unified/} and suffix
\texttt{\_GR1ArmsAndWaistFourierHands\_Env}.

\begingroup
\small
\setlength{\tabcolsep}{4pt}
\renewcommand{\arraystretch}{1.25}
\begin{longtable}{@{}p{0.20\linewidth}>{\raggedright\arraybackslash}p{0.10\linewidth}p{\dimexpr0.74\linewidth-4\tabcolsep\relax}@{}}
\caption{Task groups in RoboCasa GR1.}\label{tab:app-gr1-tasks}\\
\toprule
\textbf{Task group} & \textbf{No. of Tasks} & \textbf{Task names} \\
\midrule
\endfirsthead
\multicolumn{3}{l}{\tablename~\thetable\ (continued)}\\
\toprule
\textbf{Task group} & \textbf{Tasks} & \textbf{Task names} \\
\midrule
\endhead
\midrule
\multicolumn{3}{r}{Continued on next page}\\
\endfoot
\bottomrule
\endlastfoot
Articulated & 6 & \begin{tabular}[t]{@{}>{\ttfamily}l@{}}
PnPCupToDrawerClose \\
PnPPotatoToMicrowaveClose \\
PnPMilkToMicrowaveClose \\
PnPBottleToCabinetClose \\
PnPWineToCabinetClose \\
PnPCanToDrawerClose
\end{tabular} \\
\midrule
Tableware & 9 & \begin{tabular}[t]{@{}>{\ttfamily}l@{}}
PosttrainPnPNovelFromCuttingboardToPanSplitA \\
PosttrainPnPNovelFromCuttingboardToPotSplitA \\
PosttrainPnPNovelFromPlacematToBowlSplitA \\
PosttrainPnPNovelFromPlacematToPlateSplitA \\
PosttrainPnPNovelFromPlateToBowlSplitA \\
PosttrainPnPNovelFromPlateToPanSplitA \\
PosttrainPnPNovelFromPlateToPlateSplitA \\
PosttrainPnPNovelFromTrayToPlateSplitA \\
PosttrainPnPNovelFromTrayToPotSplitA
\end{tabular} \\
\midrule
Storage & 9 & \begin{tabular}[t]{@{}>{\ttfamily}l@{}}
PosttrainPnPNovelFromCuttingboardToBasketSplitA \\
PosttrainPnPNovelFromCuttingboardToCardboardboxSplitA \\
PosttrainPnPNovelFromCuttingboardToTieredbasketSplitA \\
PosttrainPnPNovelFromPlacematToBasketSplitA \\
PosttrainPnPNovelFromPlacematToTieredshelfSplitA \\
PosttrainPnPNovelFromPlateToCardboardboxSplitA \\
PosttrainPnPNovelFromTrayToCardboardboxSplitA \\
PosttrainPnPNovelFromTrayToTieredbasketSplitA \\
PosttrainPnPNovelFromTrayToTieredshelfSplitA
\end{tabular} \\
\end{longtable}
\endgroup

\clearpage
\subsection{Training Hyperparameters}
\label{app:training-configurations}

Tables~\ref{tab:app-libero-config} and~\ref{tab:app-robocasa-config}
report the two-stage training configurations for LIBERO and RoboCasa. At each Stage~1 iteration, we collect rollouts with the current model and take one optimizer step on the distillation loss aggregated across task groups. Stage~2 combines PPO with teacher distillation and jointly updates the value network. The coefficients of the PPO policy loss and value loss are both fixed at~1.

\begingroup
\small
\setlength{\tabcolsep}{4pt}
\renewcommand{\arraystretch}{1.30}
\begin{longtable}{@{}p{0.29\linewidth}p{\dimexpr0.41\linewidth-6\tabcolsep\relax}p{0.15\linewidth}p{0.15\linewidth}@{}}
\caption{Two-stage training configurations for LIBERO.}\label{tab:app-libero-config}\\
\toprule
\textbf{Parameter} & \textbf{Description} & \textbf{Stage 1} & \textbf{Stage 2} \\
\midrule
\endfirsthead
\multicolumn{4}{l}{\tablename~\thetable\ (continued)}\\
\toprule
\textbf{Parameter} & \textbf{Description} & \textbf{Stage 1} & \textbf{Stage 2} \\
\midrule
\endhead
\midrule
\multicolumn{4}{r}{Continued on next page}\\
\endfoot
\bottomrule
\endlastfoot
\texttt{total\_num\_envs} & Total parallel training environments & 16 & 64 \\
\texttt{suite\_num\_envs} & Parallel environments per task group $\rho$ & \shortstack[l]{[3, 3, 3, 7]} & \shortstack[l]{[13, 11, 13, 27]} \\
\texttt{task\_loss\_weights} & Task-group weights $\alpha$ for teacher distillation & \shortstack[l]{[0.25, 0.25,\\ 0.25, 0.25]} & \shortstack[l]{[0.23, 0.20,\\ 0.23, 0.34]} \\
\texttt{suite\_max\_episode\_steps} & Maximum environment steps per episode & \shortstack[l]{[220, 280, 300, 520]} & \shortstack[l]{[220, 280, 300, 520]} \\
\texttt{max\_steps\_per\_rollout\_epoch} & Rollout steps per environment & 520 & 520 \\
\texttt{action\_horizon} & Number of actions in each action chunk & 16 & 16 \\
\texttt{num\_action\_chunks} & Number of actions executed per chunk & 5 & 5 \\
\texttt{num\_steps} & Number of flow integration steps & 4 & 4 \\
\texttt{noise\_level} & Exploration noise coefficient during rollout & --- & 0.5 \\
\texttt{lr} & Model learning rate & $5\times 10^{-5}$ & $1\times 10^{-6}$ \\
\texttt{value\_lr} & Value network learning rate & --- & $1\times 10^{-4}$ \\
\texttt{micro\_batch\_size} & Microbatch size per GPU & 4 & 32 \\
\texttt{global\_batch\_size} & Global batch size per update & 1664 & 256 \\
\texttt{rollout\_epoch} & Rollout rounds per training iteration & 1 & 1 \\
\texttt{update\_epoch} & Optimization epochs over each rollout batch & 1 & 1 \\
\texttt{beta} & Teacher distillation loss coefficient $\beta$ & --- & 10.0 \\
\texttt{gamma} & Reward discount factor & --- & 0.99 \\
\texttt{gae\_lambda} & GAE parameter & --- & 0.95 \\
\texttt{clip\_ratio\_high} & Upper PPO clipping coefficient & --- & 0.2 \\
\texttt{clip\_ratio\_low} & Lower PPO clipping coefficient & --- & 0.2 \\
\texttt{value\_clip} & Value clipping coefficient & --- & 0.2 \\
\end{longtable}
\endgroup

\noindent\textit{Note.} Vector entries follow the suite order: Spatial, Object, Goal, and Long. A dash indicates that the parameter is not applicable in that stage.

\clearpage
\begingroup
\small
\setlength{\tabcolsep}{4pt}
\renewcommand{\arraystretch}{1.30}
\begin{longtable}{@{}p{0.29\linewidth}p{\dimexpr0.41\linewidth-6\tabcolsep\relax}p{0.15\linewidth}p{0.15\linewidth}@{}}
\caption{Two-stage training configurations for RoboCasa.}\label{tab:app-robocasa-config}\\
\toprule
\textbf{Parameter} & \textbf{Description} & \textbf{Stage 1} & \textbf{Stage 2} \\
\midrule
\endfirsthead
\multicolumn{4}{l}{\tablename~\thetable\ (continued)}\\
\toprule
\textbf{Parameter} & \textbf{Description} & \textbf{Stage 1} & \textbf{Stage 2} \\
\midrule
\endhead
\midrule
\multicolumn{4}{r}{Continued on next page}\\
\endfoot
\bottomrule
\endlastfoot
\texttt{total\_num\_envs} & Total parallel training environments & 24 & 72 \\
\texttt{suite\_num\_envs} & Parallel environments per task group $\rho$ & [8, 8, 8] & [24, 24, 24] \\
\texttt{task\_loss\_weights} & Task-group weights $\alpha$ for teacher distillation & [0.4, 0.4, 0.2] & [0.35, 0.35, 0.30] \\
\texttt{max\_episode\_steps} & Maximum environment steps per episode & 1000 & 1000 \\
\texttt{max\_steps\_per\_rollout\_epoch} & Rollout steps per environment & 1024 & 1024 \\
\texttt{action\_horizon} & Number of actions in each action chunk & 16 & 16 \\
\texttt{num\_action\_chunks} & Number of actions executed per chunk & 16 & 16 \\
\texttt{num\_steps} & Number of flow integration steps & 4 & 4 \\
\texttt{noise\_level} & Exploration noise coefficient during rollout & --- & 0.5 \\
\texttt{lr} & Model learning rate & $1\times 10^{-6}$ & $5\times 10^{-7}$ \\
\texttt{value\_lr} & Value network learning rate & --- & $5\times 10^{-5}$ \\
\texttt{micro\_batch\_size} & Microbatch size per GPU & 16 & 32 \\
\texttt{global\_batch\_size} & Global batch size per update & 1536 & 256 \\
\texttt{rollout\_epoch} & Rollout rounds per training iteration & 1 & 1 \\
\texttt{update\_epoch} & Optimization epochs over each rollout batch & 1 & 1 \\
\texttt{beta} & Teacher distillation loss coefficient $\beta$ & --- & 10.0 \\
\texttt{gamma} & Reward discount factor & --- & 0.99 \\
\texttt{gae\_lambda} & GAE parameter & --- & 0.95 \\
\texttt{clip\_ratio\_high} & Upper PPO clipping coefficient & --- & 0.2 \\
\texttt{clip\_ratio\_low} & Lower PPO clipping coefficient & --- & 0.2 \\
\texttt{value\_clip} & Value clipping coefficient & --- & 0.2 \\
\end{longtable}
\endgroup

\noindent\textit{Note.} Vector entries follow the order PnP,
Articulated, and Appliance--Coffee for Kitchen, and Articulated,
Tableware, and Storage for GR1. A dash indicates that the
parameter is not applicable in that stage.

\end{document}